\pdfoutput=1
\documentclass[11pt]{article}

\usepackage{lipsum}  
\usepackage{booktabs}
\usepackage{graphicx}
\usepackage{amsmath}
\usepackage{amssymb}
\usepackage{amsbsy}
\usepackage{xspace}
\usepackage{multirow}
\DeclareMathOperator*{\argmax}{argmax}
\usepackage{url}
\DeclareMathOperator*{\argmin}{arg\,min}

\newcommand\BLEU{\textsc{Bleu}\xspace}
\newcommand\TER{\textsc{Ter}\xspace}

\usepackage[]{ACL2023}

\usepackage{times}
\usepackage{latexsym}

\usepackage[T1]{fontenc}
\usepackage[utf8]{inputenc}

\usepackage{microtype}

\usepackage{inconsolata}

\title{Improving Long Context Document-Level Machine Translation}

\author{
Christian Herold \qquad Hermann Ney \\
Human Language Technology and Pattern Recognition Group \\
Computer Science Department\\
RWTH Aachen University \\
D-52056 Aachen, Germany \\
{\tt \{herold|ney\}@cs.rwth-aachen.de}
}

\begin{document}
\maketitle
\begin{abstract}
Document-level context for neural machine translation (NMT) is crucial to improve the translation consistency and cohesion, the translation of ambiguous inputs, as well as several other linguistic phenomena.
Many works have been published on the topic of document-level NMT, but most restrict the system to only local context, typically including just the one or two preceding sentences as additional information.
This might be enough to resolve some ambiguous inputs, but it is probably not sufficient to capture some document-level information like the topic or style of a conversation.
When increasing the context size beyond just the local context, there are two challenges: (i)~the~memory usage increases exponentially (ii) the translation performance starts to degrade.
We argue that the widely-used attention mechanism is responsible for both issues.
Therefore, we propose a constrained attention variant that focuses the attention on the most relevant parts of the sequence, while simultaneously reducing the memory consumption.
For evaluation, we utilize targeted test sets in combination with novel evaluation techniques to analyze the translations in regards to specific discourse-related phenomena.
We find that our approach is a good compromise between sentence-level NMT vs attending to the full context, especially in low resource scenarios.

\end{abstract}

\section{Introduction}

Machine translation (MT) is the task of mapping some input text onto the corresponding translation in the target language.
MT systems typically operate on the sentence-level and utilize neural networks trained on large amounts of bilingual data \cite{bahdanau2014neural, vaswani2017attention}.
These neural machine translation (NMT) systems perform remarkably well on many domains and language pairs, sometimes even on par with professional human translators.
However, when the automatic translations are evaluated on the document-level (e.g. the translation of a whole paragraph or conversation is evaluated), they reveal shortcomings regarding consistency in style, entity-translation or correct inference of the gender, among other things \cite{laubli2018has, muller2018large, thai2022exploring}.
The goal of document-level NMT is to resolve these shortcomings by including context information as additional input when translating a sentence.

In recent years, many works have been published on the topic of document-level NMT.
However, most of these works focus only on including a few surrounding sentences as context.
When the context size is increased beyond that, typically a degradation of overall translation performance is reported.
Additionally, the transformer architecture as the quasi standard in NMT seems sub optimal to handle long sequences as input/output, since the memory complexity increases quadratically with the sequence length.
This is due to the attention mechanism, where each token in a sequence needs to attend to all other tokens.

In this work, we propose a constrained attention variant for the task of document-level NMT.
The idea is to reduce the memory consumption while at the same time focusing the attention of the system onto the most relevant parts of the sequence.
Our contributions are two-fold:
\begin{enumerate}
    \item We observe that the attention patterns become less focused on the current sentence when increasing the context-size of our document-level NMT systems. Therefore we propose a constrained attention variant that is also more memory efficient.
    \item We utilize a targeted evaluation method to assess automatic translations in regards to consistency in style and coreference resolution. We find that our document-level NMT approach performs among the best across all language-pairs and test scenarios.
\end{enumerate}

\section{Related Work}
\label{sec:related_work}

Many works have been published on the topic of document-level NMT.
The widely used baseline approach consists of simply concatenating a few adjacent sentences and feeding this as an input to the MT system, without modifying the system architecture in any way \cite{tiedemann2017neural, bawden2017evaluating, agrawal2018contextual, talman-etal-2019-university, nguyen2021data, majumde2022baseline}.
Also, several modifications to this baseline concatenation approach have been proposed.
\citet{ma2020simple} introduce segment embeddings and also partially constrain the attention to the tokens of the current sentence. 
\citet{zhang-etal-2020-long} propose to calculate the self-attention both on the sentence- and on the document-level and then combine the two representations.
\citet{fernandes2021measuring} and \citet{lei-etal-2022-codonmt} both mask out tokens in the current sentence to increase context utilization while \citet{yang2023hanoit} remove tokens from the context if they are not attended.
Typically, slight improvements in \BLEU are reported as well as more significant improvements on targeted test sets e.g. for coreference resolution. 

Apart from the simple concatenation method, there exist other approaches to document-level NMT.
They include using a single document-embedding vector \cite{mace-servan-2019-using, stojanovski-fraser-2019-combining, jiang2020document, huo2020diving}, multiple encoders \cite{jean2017does, bawden2017evaluating, wang2017exploiting, voita2018context, zhang2018improving}, hierarchical attention \cite{miculicich2018document, maruf2019selective, tan-etal-2019-hierarchical, wong2020contextual}, translation caches \cite{maruf2018document, tu2018learning, kuang-etal-2018-modeling} or dynamic evaluation \cite{mansimov-etal-2021-capturing}.
However, these approaches are less versatile and require significant changes to the model architecture, often introducing a significant amount of additional parameters.
Furthermore, recent works have concluded that the baseline concatenation approach first proposed by \citet{tiedemann2017neural} performs as good - if not better - than these more complicated approaches \cite{lopes2020document, sun2022rethinking}.

While the concatenation approach works well for short context sizes, when used with a larger number of context sentences, typically performance degradation is reported:
\citet{scherrer-etal-2019-analysing} saw a severe performance degradation when using input sequences with a length of 250 tokens.
\citet{10.1162/tacl_a_00343} could not get their system to converge when using context sizes of up to 512 tokens.
They improve training stability by adding additional monolingual data via pre-training.
\citet{bao2021g} also report that their systems with context length of more than 256 tokens fail to converge.
They propose to partially constrain the attention to the current sentence, similar to \citet{zhang-etal-2020-long}.
\citet{sun2022rethinking} try to translate full documents with the concatenation approach but could not get their system to converge during training.
Their solution is to mix document- and sentence-level data, which reportedly improves system convergence.
\citet{li2022p} report severe performance degradation for context sizes longer than 512 tokens.
They argue this is due to insufficient positional information and improve performance by repeatedly injecting this information during the encoding process.
However, increasing the context size seems to not always result in performance degradation.
In their works, \citet{junczys-dowmunt-2019-microsoft} and \citet{saleh-etal-2019-naver} train systems with a context size of up to 1000 tokens without degradation in translation quality, which stands in contrast to the works mentioned above and which we will discuss again in the context of our own results.
We want to point out that all of the approaches mentioned above still have the problem of quadratically increasing resource requirements, which poses a big challenge even on modern hardware.

Since our proposed approach consists of modifying the attention matrix in the model architecture, we give a brief overview of previous works related to this concept.
The works of \citet{ma2020simple}, \citet{zhang-etal-2020-long} and \citet{bao2021g}  are most closely related and were already mentioned above.
All three papers restrict the attention (partially) to the current sentence and combine sentence- and document-level attention context vectors for the final output.
However, this means all approaches still suffer from the quadratic dependency on the number of input tokens.
\citet{luong-etal-2015-effective} were among the first to propose using the attention concept for the task of MT.
They also proposed using a sliding-window with target-to-source alignment for attention similar to us.
However, they only work on sentence-level NMT and to the best of our knowledge, this approach was never before transferred to document-level NMT.
\citet{shu2017empirical} and \citet{Chen_Wang_Utiyama_Sumita_Zhao_2018} both extend the approach of \citet{luong-etal-2015-effective} while still working solely on sentence-level NMT.
Our approach is also related to the utilization of relative positional encoding, which was introduced by \citet{shaw-etal-2018-self} and later extended by \citet{yang-etal-2018-modeling} to be applicable for cross-attention.
The work by \citet{indurthi-etal-2019-look} should also be mentioned, where they pre-select a subset of source tokens on which to perform attention on.
Again, all of the above mentioned works only perform experiments on sentence-level NMT.
The works of \citet{child2019generating}, \citet{sukhbaatar-etal-2019-adaptive} and \citet{Guo_Qiu_Liu_Xue_Zhang_2020} are also related, since they use attention windows similar to us for tasks other than MT.

Finally, we briefly want to touch on the subject of automatic evaluation of document-level MT systems.
Many works only report results on general MT metrics like \BLEU \cite{papineni2002bleu}, sometimes matching n-grams across sentence-boundaries.
However, it has been argued that these metrics do not capture well the very specific improvements that could be expected by including document-level context and that the reported improvements rather come from regularization effects and comparing to sub optimal baseline performance \cite{kim2019and, li2020does, nguyen2021data}.
Several targeted test suites have been released to better assess the improvements gained by document-level NMT \cite{muller2018large, bawden2017evaluating, voita2019good, jwalapuram2019evaluating}.
These test suites have some limitations, for example they are language-specific and they are based on just scoring predefined contrastive examples without scoring the actual translations.
More recently, \citet{jiang-etal-2022-blonde} and \citet{currey2022mt} have released frameworks that allow to score the actual MT hypotheses in regards to their consistency regarding specific aspects of the translation.

\section{Methodology}
\label{sec:methodology}

Here, we explain the baseline concatenation approach (Section \ref{subsec:concat}), the more refined method that we are comparing ourselves against (Section \ref{subsec:lst}) as well as our own approach (Section \ref{subsec:window_mask}).
We also discuss our different evaluation approaches in Section \ref{subsec:eval}.

\subsection{The Baseline Concatenation Approach}
\label{subsec:concat}

The baseline concatenation approach is very simple and follows \citet{tiedemann2017neural} using the vanilla transformer architecture \cite{vaswani2017attention}.
Assume we are given a document $\mathcal{D}=(F_n, E_n)_1^N$ consisting of $N$ source-target sentence-pairs $(F_n, E_n)$.
If we want our model to have a context length of $k$ sentences, we simply concatenate the current input sentence with its $k-1$ predecessor sentences and the input to the model would be
\begin{center}
    $F_{n-k}$ \texttt{<sep>} $F_{n-k+1}$ ... \texttt{<sep>} $F_{n}$ \texttt{<eos>}
\end{center}
while on the target side we include the preceding sentences as a prefix
\begin{center}
    $E_{n-k}$ \texttt{<sep>} $E_{n-k+1}$ ... \texttt{<sep>} $E_{n-1}$ \texttt{<sep>}. \\
\end{center}
We use a special token \texttt{<sep>} as a separator between adjacent sentences and \texttt{<eos>} denotes the end of the sequence.
This is done to make it easier for the model to distinguish between the sentence that needs to be translated and the context.
Furthermore, we use a special token $F_0=E_0=$ \texttt{<bod>} to denote the start of a document.
Since we use the vanilla transformer architecture with self-attention and cross-attention components, the memory usage is $\mathcal{O}(L^2)$ with $L$ being the sequence length.

When we train full document-level systems, we simply concatenate all sentences in the document using again the special \texttt{<sep>} token.
Due to hardware limitations, if the length of the target-side of the document exceeds 1000 tokens, we split the document into smaller parts of roughly equal length (i.e. a document of length 1500 tokens would be split into two parts with ca. 750 tokens each).

\begin{table}[]
\centering
\begin{tabular}{l|r|r|r}
\toprule
\multicolumn{1}{c|}{Model} & \multicolumn{1}{c|}{Context}                        & \multicolumn{1}{c|}{Attn. {[}\%{]}} & \multicolumn{1}{c}{\BLEU} \\ 
 \hline
sent.-level     & \multicolumn{1}{r|}{0\,sent.}      & 100.0 & 32.8  \\ \hline
concat adj.     & \multicolumn{1}{r|}{1\,sent.}      & 76.0 & 33.1  \\
                   % & \multicolumn{1}{r|}{2s}   & 2s   &  & 33.4 & 48.6 \\
                   % & \multicolumn{1}{r|}{4s}   & 4s   &  & 33.3 & 48.8 \\
                   % & \multicolumn{1}{r|}{8s}   & 8s   &  & 32.1 & 49.9 \\ 
 & \multicolumn{1}{r|}{1000\,tok.}  & 46.6 & 29.5  \\ \bottomrule
\end{tabular}
\caption{Percentage of attention on the $n$-th source sentence during decoding the $n$-th target sentence, as well as overall translation quality measured in \BLEU, for the \texttt{newstest2018} test set of the \textit{NEWS} task.}
\label{tab:attn_percent}
\end{table}

In a preliminary study, we train systems using no context (sentence-level), just a single sentence as context as well as the maximum context size of 1000 tokens.
When looking at the percentage of attention that is payed to the $n$-th source sentence $F_n$ when decoding the $n$-th target sentence $E_n$ (extracted from cross-attention module, see Table \ref{tab:attn_percent}) we find that this percentage becomes lower as the context size increases.
This finding motivates us to explore approaches that bias the attention towards the current sentence.

\subsection{LST-attention}
\label{subsec:lst}

This method was proposed by \citet{zhang-etal-2020-long} and is called Long-Short Term (LST) attention.
The authors find that their approach outperforms the baseline concatenation approach but they only use a maximum of 3 sentences as context.
Nevertheless we deem this approach promising, since it also focuses the attention onto the current sentence.
The input to the system is augmented in the same way as described in Section \ref{subsec:concat}.
Given some queries $Q\in \mathbb{R}^{I\times d}$, keys $K\in \mathbb{R}^{J\times d}$ and values $V\in \mathbb{R}^{J\times d}$, \citet{zhang-etal-2020-long} formulate their restricted version of the attention as\footnote{In practice we use multi-head attention in all our architectures, but we omit this in the formulas for sake of simplicity. Also, for all methods, causal masking is applied in the decoder self-attention just like in the vanilla transformer.}
\begin{equation*}
    \text{A}(Q, K, V) = \text{softmax}\left(\frac{Q \cdot K^\intercal}{\sqrt{d}} + M\right)V
\end{equation*}
with $d$ being the hidden dimension of the model and $M\in \mathbb{R}^{I\times J}$ being the masking matrix.
This masking matrix is defined as 
\begin{equation*}
  M_{i, j} =\ \begin{cases}
    0 & , s(i) = s(j)\\
    -\inf & ,\text{otherwise}
  \end{cases}
  \label{eq:def_mask_sent}
\end{equation*}
where $s(\cdot) \in 1, .., N$ is a function that returns the sentence index that a certain position belongs to.
This means we are restricting the attention to be calculated only within the current sentence.
For self-attention in the encoder and the decoder, \citet{zhang-etal-2020-long} calculate both the restricted and the non-restricted variant and then combine the output context-vectors via concatenation and a linear transformation. 
The cross-attention between encoder and decoder remains unchanged in this approach and the memory consumption remains $\mathcal{O}(L^2)$.

\subsection{window-attention}
\label{subsec:window_mask}
This method is proposed by us.
We can use the same formulation as above to describe this approach by simply changing the definition of the attention mask to
\begin{equation}
  M_{i, j} =\ \begin{cases}
    0 & , b_i - w \leq j \leq b_i + w\\
    -\inf & ,\text{otherwise}
  \end{cases}
  \label{eq:def_mask_window}
\end{equation}
where $w$ is the window size and $b_i \in 1, ..., J$ is a target-source alignment.
This means a certain query vector $q_i$ is only allowed to attend to the key vectors $k_j$ that surround the position $b_i$ that this query vector is aligned to.
We replace all self-attention and cross-attention modules in our network with this window-attention variant. 
Please note that in practice we do not calculate this mask, but instead we first select the corresponding key-vectors for each query and then calculate attention only between these subsets which reduces the memory consumption from $\mathcal{O}(L^2)$ to $\mathcal{O}(L \cdot w)$.
We also want to point out that with this approach, the context is not as restricted as it seems on first glance.
For any individual attention module, the context is restricted to $2\cdot w$ or $w$ for self-attention in the encoder and decoder respectively.
However, since in the transformer architecture we stack multiple layers, we get a final  effective context size of $2\cdot w \cdot num\_enc\_layers + w \cdot num\_dec\_layers$.

This approach requires us to define an alignment function $b_i: [1, I] \to [1, J]$.
For self-attention, we assume a 1-1 alignment so the alignment function is the identity function $b_i = i$.
For cross-attention, during training we use a linear alignment function
\begin{equation*}
    b_i = \text{round}(\frac{J}{I}\cdot i)
\end{equation*}
where $J$ is the number of tokens in the source document and $I$ is the number of tokens in the target document.
This is not possible during decoding, as we do not know the target document length beforehand.
Therefore, we propose three different ways to approximate the alignment during decoding:
\begin{enumerate}
    \item 1-1 alignment: $b_i = i$
    \item linear alignment: $b_i = \text{round}(train\_ratio \cdot i)$ where we define $train\_ratio$ as the average source-target ratio over all documents in the training data.
    \item sent-align: assume we have already produced $N'$ full target sentences (i.e. we have produced $N'$ \texttt{<sep>} tokens) up to this point, then
        \begin{equation*}
          b_i =\ \begin{cases}
            \sum_{n=1}^{N'} J_n + 1 & , e_{i-1} == \texttt{<sep>}\\
            b_{i-1} + 1 & ,\text{otherwise}
          \end{cases}
          \label{eq:def_mask_sent}
        \end{equation*}
        with $J_n$ being the length of the $n$-th source sentence in the input document.
        In simple terms, when starting to decode a new sentence, we always force-align to the beginning of the corresponding source sentence.
\end{enumerate}

We also test the \textit{window-attention} approach with relative positional encoding in the self-attention instead of absolute positional encoding, which in this framework only requires a small modification to Equation \ref{eq:def_mask_window}:
\begin{equation*}
  M_{i, j} =\ \begin{cases}
    r_{i-j} & , b_i - w \leq j \leq b_i + w\\
    -\inf & ,\text{otherwise}
  \end{cases}
\end{equation*}
where $r_{i-j} \in \mathbb{R}^{d}$ are additional learnable parameters of the network.

\subsection{Decoding}
\label{subsec:decoding}

During decoding, given a document $F_1^N$, we want to find the best translation $\hat{E}_1^N$ according to our model.
We can not perform exact search due to computational limitations, therefore we have to use approximations.
There exist multiple approaches for decoding with a document-level NMT system and since we could not determine a single best approach from the literature, we describe and compare two competing approaches. 
\begin{description}
    \item[\textit{Full Segment Decoding (FSD)}]\cite{10.1162/tacl_a_00343, bao2021g, sun2022rethinking}: we split the document into non-overlapping parts $F_1^{k}, F_{k+1}^{2k}, ..., F_{N-k}^{N}$ and translate each part separately using
    \begin{equation*}
        \hat{E}_{i-k}^{i} = \argmax_{E_{i-k}^{i}}\left\{p(E_{i-k}^{i}|F_{i-k}^{i}) \right\},
        \label{eq:1}
    \end{equation*}
    which is approximated using standard beam search on the token level (we use beam size 12 for all experiments).
    For the full document-level systems, we simply use
    \begin{equation*}
        \hat{E}_1^N = \argmax_{E_1^N}\left\{p(E_1^N|F_1^N) \right\}.
    \end{equation*}
    \item[\textit{Sequential Decoding (SD)}]\cite{miculicich2018document, voita2019good, garcia2019context, fernandes2021measuring}: we generate the translation sentence by sentence, using the previously generated target sentences as context:
    \begin{equation*}
        \hat{E}_{i} = \argmax_{E_{i}}\left\{p(E_{i}|\hat{E}_{i-k}^{i-1}, F_{i-k}^{i}) \right\}.
    \end{equation*}
\end{description}

\subsection{Evaluation}
\label{subsec:eval}

For all tasks we report \BLEU \cite{papineni2002bleu} and \TER \cite{snover2006study} using the SacreBLEU \cite{post2018call} toolkit.
In addition, for the two En-De tasks (\textit{NEWS} and \textit{OS}) we analyze the translations in regards to ambiguous pronouns and style.
For pronouns, the goal is to measure how well a system can translate the English 3rd person pronoun \lq{}it\rq{} (and its other forms) into the correctly gendered German form (which can be male, female or neuter depending on the context).
For style, the goal is to measure, how well a system can translate the 2nd person pronoun \lq{}you\rq{} (and its other forms) into the correct style in German.
For example, \lq{}you\rq{} (singular) can be translated into \lq{}sie\rq{} or \lq{}du\rq{} in German, depending if the setting is formal or informal.
We employ several strategies to determine the systems ability to disambiguate these phenomena.

We utilize the ContraPro test suite \cite{muller2018large} and report the contrastive scoring accuracy for pronoun resolution.
The test suite contains 12,000 English sentences, each with the correct German reference as well as 2 contrastive German references where the pronoun has been changed to a wrong gender.
We score all test cases with the NMT system and each time the system gives the best score to the true reference, it gets a point.
In the end we report the scoring accuracy, i.e. the number of points the system has gathered divided by 12,000.

Additionally we also report F1 scores for pronoun and style translation, the method for which is inspired by \citet{jiang-etal-2022-blonde}.
We use parts of speech (POS) taggers as well as language-specific regular expressions to identify ambiguous pronouns/formality in the test sets.
We then compare the occurrences in the reference against the occurrences in the hypothesis, calculate precision and recall and then finally the F1 score for both pronoun as well as formality translation.
The exact algorithm as well as detailed data statistics for the test sets are given in Appendix \ref{subsec:pronoun_formality_eval}.
We report pronoun translation F1 score for both \textit{NEWS} and \textit{OS} tasks and the formality translation F1 score only for the \textit{OS} task, since in the \textit{NEWS} test set there are not enough examples of ambiguous formality cases.
Our extension to the work of \citet{jiang-etal-2022-blonde} can be found here: \url{https://github.com/christian3141/BlonDe}.

\section{Experiments}
\label{sec:experiments}

We perform experiments on three document-level translation benchmarks.
We call them \textbf{NEWS} (En$\to$De) with \texttt{newstest2018} as test set, \textbf{TED} (En$\to$It) with \texttt{tst2017} as test set and \textbf{OS} (En$\to$De) where the test set is simply called \texttt{test}.
\textit{NEWS} is a collection of news articles, \textit{TED} is a collection of transcribed TED talks and their respective translations and \textit{OS} consists of subtitles for movies and TV shows.
Especially the latter holds many examples for discourse between different entities.
For the details regarding data conditions, preparation and training, we refer to Appendix \ref{subsec:data_stats}.

\subsection{GPU Memory efficiency}

First, we compare the GPU memory consumption of the baseline \textit{concat-adj.} approach against the \textit{window-attention} approach for various input sequence lengths.
The results are shown in Table \ref{tab:memory_consumption}.
\begin{table}[ht]
\centering
\begin{tabular}{l|r|rr}
\toprule
\multicolumn{1}{c|}{\multirow{2}{*}{\begin{tabular}[c]{@{}c@{}}\# target\\ tokens\end{tabular}}} & \multicolumn{1}{c|}{\multirow{2}{*}{concat-adj.}} & \multicolumn{2}{c}{window-attn} \\ \cline{3-4} 
\multicolumn{1}{c|}{} & \multicolumn{1}{c|}{} & \multicolumn{1}{c|}{$w=10$} & \multicolumn{1}{c}{$w=20$} \\ \hline
736 & 2.3\,GB & \multicolumn{1}{r|}{2.4\,GB} & 3.5\,GB \\
1472 & 5.8\,GB & \multicolumn{1}{r|}{3.9\,GB} & 5.9\,GB \\
2208 & 10.9\,GB & \multicolumn{1}{r|}{5.2\,GB} & 8.5\,GB \\ \bottomrule
\end{tabular}
\caption{GPU-memory consumption for the different approaches when training on a single document of specified number of target tokens.}
\label{tab:memory_consumption}
\end{table}
As expected, the memory usage increases at a much higher rate for the \textit{concat-adj.} approach, while the \textit{window-attention} approach scales roughly linearly, the slope being a function of the window-size $w$.

\subsection{Comparison of Decoding Strategies}

After training all models on the \textit{NEWS} task according to Appendix \ref{subsec:data_stats}, we test the different search strategies for each of the systems, the result of which can be found in Table \ref{tab:search_experiments}.
\begin{table}[]
\centering
\begin{tabular}{l|l|l|r}
\toprule
\multicolumn{1}{c|}{Model} & \multicolumn{1}{c|}{Context} & \multicolumn{1}{c|}{\begin{tabular}[c]{@{}c@{}}Search\\ Strategy\end{tabular}} & \multicolumn{1}{c}{\BLEU} \\ \hline
sent.-level & 0 sent. & - & 32.8 \\ \hline
concat-adj. & 2 sent. & \textit{FSD} & 33.4 \\ \cline{3-4} 
 &  & \textit{SD} & 33.0 \\ \cline{2-4} 
 & 1000 tok. & \textit{FSD} & 29.5 \\ \cline{3-4} 
 &  & \textit{SD} & 23.1 \\
 \hline
LST-attn & 1000 tok. & \textit{FSD} & 30.0 \\ \cline{3-4} 
 &  & \textit{SD} & 22.2 \\
 \hline
window-attn & 1000 tok. & \textit{FSD} & 31.5 \\ \cline{3-4} 
 &  & \textit{SD} & 33.1 \\
 \bottomrule
\end{tabular}
\caption{Results for employing the different search strategies for translating the \texttt{newstest2018} test set of the \textit{NEWS} task.}
\label{tab:search_experiments}
\end{table}
For the baseline \textit{concat-adj.} approach as well as the \textit{LST-attn} approach, \textit{FSD} works best.
However, we still see significant performance degradation for the systems using long context information.
For \textit{concat-adj.} and \textit{LST-attn} with 1000 tokens context size, \textit{SD} performs very poorly.
This is because when beginning translating a document, the input sequences are very short and the systems can not appropriately handle that.
However, \textit{FSD} sometimes leads to sentence-misalignment while translating a document, resulting in a lower \BLEU score as well.
For the \textit{window-attention} approach (rel. pos. enc., sent-align, window-size 20) we find that the \textit{SD} decoding strategy works best.
Since this approach seems to be able to better handle short input sequences, \textit{SD} performs better than \textit{FSD}, since it seems more robust to sentence-misalignment.
Moving forward, all numbers reported will be generated with the best respective decoding approach, i.e. \textit{SD} for \textit{window-attention} and \textit{FSD} for all other approaches.

\subsection{Hyperparameter Tuning}

Our \textit{window-attention} approach has three hyperparameters that need to be tuned: (i) positional encoding variant (ii) alignment variant during search (iii) window size.
Again, we use the \textit{NEWS} task for tuning and the results for the different variants can be found in Table \ref{tab:hyperparams}.
\begin{table*}[t!]
\centering
\begin{tabular}{l|r|r|r|r|r}
\toprule
\multicolumn{1}{c|}{Model} & \multicolumn{1}{c|}{pos. enc.} & \multicolumn{1}{c|}{Alignment} & \multicolumn{1}{c|}{window-size} & \multicolumn{1}{c|}{\BLEU} & \multicolumn{1}{c}{\TER} \\ \hline
concat-adj. & abs. & - & - & 29.5 & 53.7 \\ \cline{2-6} 
 & rel. & - & - & N/A & N/A \\ \hline
window-attn & abs. & 1-1 & 20 & 29.7 & 51.7 \\ \cline{3-6} 
 &  & train avg. & 20 & 28.1 & 55.3 \\ \cline{3-6} 
 &  & sent-align & 10 & 28.3 & 53.7 \\
 &  &  & 20 & 30.3 & 50.9 \\
 &  &  & 30 & 29.4 & 52.2 \\ \cline{2-6} 
 & rel. & 1-1 & 20 & 31.9 & 49.8 \\ \cline{3-6} 
 &  & train avg. & 20 & 30.5 & 53.2 \\ \cline{3-6} 
 &  & sent-align & 10 & 30.6 & 51.8 \\
 & \multicolumn{1}{l|}{} &  & 20 & 33.1 & 48.1 \\
 & \multicolumn{1}{l|}{} &  & 30 & 32.8 & 48.4 \\ \bottomrule
\end{tabular}
\caption{Results for the different hyperparameter settings of the \textit{window-attention} system reported on the \texttt{newstest2018} test set of the \textit{NEWS} task. All systems have context size 1000 tokens.}
\label{tab:hyperparams}
% \end{table*}
\vspace{0.5cm}
% \begin{table*}[]
% \centering
\begin{tabular}{l|l|rr|rr|rr}
\toprule
\multicolumn{1}{c|}{\multirow{3}{*}{Model}} & \multicolumn{1}{c|}{\multirow{3}{*}{Context}} & \multicolumn{2}{c|}{NEWS} & \multicolumn{2}{c|}{TED} & \multicolumn{2}{c}{OS} \\ \cline{3-8} 
 &  & \multicolumn{2}{c|}{newstest2018} & \multicolumn{2}{c|}{tst2017} & \multicolumn{2}{c}{test} \\ \cline{3-8} 
 &  & \multicolumn{1}{c|}{\BLEU} & \multicolumn{1}{c|}{\TER} & \multicolumn{1}{c|}{\BLEU} & \multicolumn{1}{c|}{\TER} & \multicolumn{1}{c|}{\BLEU} & \multicolumn{1}{c}{\TER} \\ \hline
sent.-level (external) & 0 sent. & \multicolumn{1}{r|}{$^{\dagger}$32.3} & - & \multicolumn{1}{r|}{$^{\ddagger}$33.4} & - & \multicolumn{1}{r|}{*37.3} & - \\
sent.-level (ours) &  & \multicolumn{1}{r|}{32.8} & 49.0 & \multicolumn{1}{r|}{34.2} & 46.3 & \multicolumn{1}{r|}{37.1} & \textbf{43.8} \\ \hline
concat adj. & 2 sent. & \multicolumn{1}{r|}{\textbf{33.4}} & 48.6 & \multicolumn{1}{r|}{34.3} & 46.3 & \multicolumn{1}{r|}{38.2} & 43.9 \\
 & 1000 tok. & \multicolumn{1}{r|}{29.5} & 53.7 & \multicolumn{1}{r|}{32.1} & 48.4 & \multicolumn{1}{r|}{38.1} & 46.0 \\ \hline
LST-attn & 1000 tok. & \multicolumn{1}{r|}{30.0} & 53.1 & \multicolumn{1}{r|}{29.8} & 54.5 & \multicolumn{1}{r|}{\textbf{38.5}} & 45.1 \\ \hline
window-attn & 1000 tok. & \multicolumn{1}{r|}{33.1} & \multicolumn{1}{l|}{\textbf{48.1}} & \multicolumn{1}{r|}{\textbf{34.6}} & \textbf{45.8} & \multicolumn{1}{r|}{38.3} & 44.4 \\ \bottomrule
\end{tabular}
\caption{Results for the different document-level approaches in terms of \BLEU and \TER on the three translation benchmarks. Best results for each column are highlighted. External baselines are from $^{\dagger}$ \citet{kim2019and}, $^{\ddagger}$ \citet{yang2022gtrans} and *\citet{huo2020diving}.}
\label{tab:final_bleu}
\end{table*}

In terms of positional encoding, \textit{relative} works significantly better than \textit{absolute} for the \textit{window-attention} system.
We also test relative positional encoding (window-size 20) for the baseline \textit{concat-adj.} method, but here the training did not converge.
This is, because for long input sequences the system without explicit target-source alignment can no longer distinguish the token ordering on the source side (on the target side it is still possible due to the causal attention mask).
The only way to resolve this would be to drastically increase the window-size for the relative positions, however, this would add a significant amount of additional parameters to the network so we decide against this.
In terms of alignment, using the \textit{sent-align} variant significantly outperforms the other approaches.
For the \textit{window-size}, 20 works best.
An important finding is, that if we make the window too large, we start losing performance, probably due to the less focused attention problem discussed in Section \ref{subsec:concat}.

% moved due to formatting reasons
\begin{table*}[t!]
\centering
\begin{tabular}{l|l|rrr|rrrr}
\toprule
\multicolumn{1}{c|}{\multirow{3}{*}{Model}} & \multicolumn{1}{c|}{\multirow{3}{*}{Context}} & \multicolumn{3}{c|}{NEWS} & \multicolumn{4}{c}{OS} \\ \cline{3-9} 
\multicolumn{1}{c|}{} & \multicolumn{1}{c|}{} & \multicolumn{3}{c|}{ContraPro} & \multicolumn{3}{c|}{ContraPro} & \multicolumn{1}{c}{test} \\ \cline{3-9} 
\multicolumn{1}{c|}{} & \multicolumn{1}{c|}{} & \multicolumn{1}{c|}{\BLEU} & \multicolumn{1}{c|}{\begin{tabular}[c]{@{}c@{}}Scoring\\ Acc.\\ Pronoun\end{tabular}} & \multicolumn{1}{c|}{\begin{tabular}[c]{@{}c@{}}Pronoun\\ Trans.\\ F1\end{tabular}} & \multicolumn{1}{c|}{\BLEU} & \multicolumn{1}{c|}{\begin{tabular}[c]{@{}c@{}}Scoring\\ Acc.\\ Pronoun\end{tabular}} & \multicolumn{1}{c|}{\begin{tabular}[c]{@{}c@{}}Pronoun\\ Trans.\\ F1\end{tabular}} & \multicolumn{1}{c}{\begin{tabular}[c]{@{}c@{}}Formality\\ Trans.\\ F1\end{tabular}} \\ \hline
sent.-level & 0 sent. & \multicolumn{1}{r|}{18.4} & \multicolumn{1}{r|}{48.2} & 44.5 & \multicolumn{1}{r|}{29.7} & \multicolumn{1}{r|}{45.8} & \multicolumn{1}{r|}{40.3} & 59.4 \\ \hline
concat adj. & 2 sent. & \multicolumn{1}{r|}{\textbf{19.6}} & \multicolumn{1}{r|}{\textbf{67.9}} & \textbf{54.1} & \multicolumn{1}{r|}{31.2} & \multicolumn{1}{r|}{81.8} & \multicolumn{1}{r|}{63.2} & 61.7 \\
 & 1000 tok. & \multicolumn{1}{r|}{15.4} & \multicolumn{1}{r|}{61.9} & 47.8 & \multicolumn{1}{r|}{29.9} & \multicolumn{1}{r|}{83.1} & \multicolumn{1}{r|}{64.6} & 70.1 \\ \hline
LST-attn & 1000 tok. & \multicolumn{1}{r|}{16.8} & \multicolumn{1}{r|}{61.4} & 51.3 & \multicolumn{1}{r|}{29.1} & \multicolumn{1}{r|}{83.3} & \multicolumn{1}{r|}{64.8} & \textbf{70.9} \\ \hline
window-attn & 1000 tok. & \multicolumn{1}{r|}{\textbf{19.6}} & \multicolumn{1}{r|}{63.0} & 51.9 & \multicolumn{1}{r|}{\textbf{31.4}} & \multicolumn{1}{r|}{\textbf{83.9}} & \multicolumn{1}{r|}{\textbf{66.5}} & 67.9 \\ \bottomrule
\end{tabular}
\caption{Results for the different document-level approaches in terms of pronoun and formality translation. Best results for each column are highlighted.}
\label{tab:final_pro}

\vspace{0.5cm}

\makebox[0cm]{
\resizebox{16cm}{!}{
\begin{tabular}{ll}
 \multicolumn{1}{c}{source} & \multicolumn{1}{c}{reference} \\ \hline
What's between you and Dr. Webber - is none of my business... & Was zwischen \textcolor{blue}{dir} und Dr. Webber ist, geht mich nichts an... \\ 
- You don't owe me an apology. & \textcolor{blue}{Du} schuldest mir keine Entschuldigung. \\ 
You owe Dr. Bailey one. & \textcolor{blue}{Du} schuldest Dr. Bailey eine.  \\ 
We were taking a stand for Dr. Webber. & Wir haben uns für Dr. Webber eingesetzt. \\ 
I don't understand why... & Ich verstehe nicht wieso... \\ 
Dr. Webber doesn't need you to fight his battles. & Dr. Webber braucht \textcolor{blue}{dich} nicht, um seine Schlachten zu kämpfen. \\ 
What you did stands to hurt this entire hospital. & Was \textcolor{blue}{du} getan hast, hat dem ganzen Krankenhaus geschadet.  \\ 
Your first priority needs to be this place and its patients. & \textcolor{blue}{Deine} oberste Priorität muss diesem Haus und seinen Patienten gelten. \\ 
 & \\
 \multicolumn{1}{c}{sentence-level-hypothesis} & \multicolumn{1}{c}{window-mask-hypothesis} \\ \hline
Was zwischen \textcolor{red}{Ihnen} und Dr. Webber ist, geht mich nichts an... & Was zwischen \textcolor{blue}{dir} und Dr. Webber ist, geht mich nichts an... \\ 
- \textcolor{blue}{Du} schuldest mir keine Entschuldigung. & - \textcolor{blue}{Du} schuldest mir keine Entschuldigung. \\ 
\textcolor{red}{Sie} schulden Dr. Bailey etwas. & - \textcolor{blue}{Du} schuldest Dr. Bailey eine.  \\ 
Wir haben für Dr. Webber Partei ergriffen. & Wir haben für Dr. Webber Stellung bezogen. \\ 
Ich verstehe nicht, warum... & Ich verstehe nicht, warum... \\ 
Dr. Webber braucht \textcolor{red}{Sie} nicht, um seine Schlachten zu schlagen. & Dr. Webber braucht \textcolor{blue}{dich} nicht, um seine Schlachten zu kämpfen. \\ 
Was \textcolor{blue}{du} getan hast, verletzt das gesamte Krankenhaus. & Was \textcolor{blue}{du} getan hast, könnte das ganze Krankenhaus verletzen.  \\ 
\textcolor{red}{Ihre} oberste Priorität muss dieser Ort und seine Patienten sein. & \textcolor{blue}{Deine} oberste Priorität muss dieser Ort und seine Patienten sein. \\ 
\end{tabular}
}
}

\caption{Example translation of a snippet from the OpenSubtitles \texttt{test} set. Formal 2nd person pronouns are marked in \textcolor{red}{red} and informal ones are marked in \textcolor{blue}{blue}.}
\label{tab:example_formality}

\end{table*}

\subsection{Final Performance Comparison}

In Table \ref{tab:final_bleu} we report the translation performance of the different document-level approaches on all three translation benchmarks measured in terms of \BLEU and \TER. 
None of the document-level systems can consistently outperform the sentence-level baseline on all tasks.
On the \textit{OS} test set, there is a disagreement between \BLEU and \TER which we think comes from the fact that the average sentence-length on this test set is quite short. 
The hypothesis of the sentence-level system is the shortest of all hypotheses and also shorter than the reference which gets punished more heavily by \BLEU than \TER.
Out of all full-document approaches, \textit{window-attention} performs best and is on par with the sentence-level baseline and the document-level system using only 2 sentences as context.
For full-document translation, \textit{LST-attn} performs better than the baseline concatenation approach but still falls behind the sentence-level system especially on the \textit{NEWS} and \textit{TED} tasks.
One possible reason for why these approaches work better on \textit{OS} is, that for this task we have much more training data available than for \textit{NEWS} and \textit{TED}.
We argue that this could also be the reason for the conflicting results reported by \citet{junczys-dowmunt-2019-microsoft} and \citet{saleh-etal-2019-naver} compared to the other works who report performance degradation for longer context sizes (see Section \ref{sec:related_work}).
However, we leave a detailed analysis of this for future work.

Next, we analyze the ability of the systems to translate ambiguous pronouns and to translate in a consistent style using the methods explained in Section \ref{subsec:eval}.
The results for the two En$\to$De tasks can be found in Table \ref{tab:final_pro}.
For both \textit{NEWS} and \textit{OS}, all document-level systems can significantly improve over the sentence-level baseline in terms of pronoun translation.
We also find that a context longer than two sentences does not seems to help for the pronoun task.
This is actually to be expected since typically the distance between noun and pronoun is not that large and according to \citet{muller2018large}, the overwhelming majority of ContraPro test cases do not require more than two sentences as context.
For the correct translation of the style however, the larger context size is clearly beneficial, as the system with just 2 sentences as context can barely outperform the sentence-level baseline.
To correctly infer the style of a conversation, ideally the whole dialog should be part of the context, especially the beginning of the conversation.
In Table \ref{tab:example_formality}, we show a snippet of the \texttt{test} set of the \textit{OS task} together with the translations of the sentence-level system and the \textit{window-attention} system.
This example highlights the need for long-context NMT systems especially for the task of dialogue translation, since there we need to stay consistent in terms of style, which the sentence-level system can not manage.
Overall, the \textit{LST-attn} approach performs best for the task of formality translation, but the other full-document systems are not far behind.

\section{Conclusion}

In this work, we focus on methods to increase the context-size for document-level NMT systems.
We point out the shortcomings of the baseline approaches to long-context document-level NMT and in turn propose to modify the attention component to be more focused and also to be more memory efficient.
We compare our approach against approaches from literature on multiple translation tasks and using different targeted evaluation methods.
We confirm the improved memory efficiency of the proposed method.
We find that for some discourse phenomena like pronoun translation, the longer context information is not necessary.
For other aspects, like consistent style translation, the longer context is very beneficial.
It seems that the baseline concatenation approach needs large amounts of training data to perform well for larger context sizes.
We conclude that our approach performs among the best across all tasks and evaluation methods, with the additional benefit of reduced memory consumption for long input sequences.

\section*{Acknowledgements}
This work was partially supported by the project HYKIST funded by the German Federal Ministry of Health on the basis of a decision of the German Federal Parliament (Bundestag) under funding ID ZMVI1-2520DAT04A, and by NeuroSys which, as part of the initiative “Clusters4Future”, is funded by the Federal Ministry of Education and Research BMBF (03ZU1106DA).

\section*{Limitations}

This work is about document-level NMT, we focus specifically on methods that improve the model performance for long input sequences.
Due to constrained resources, this work has several limitations.
To be able to train all methods including the inefficient baseline approach, we have to limit the context size to 1000 tokens.
While we do a comparison to existing approaches, other approaches have been proposed to improve the performance of systems with long context information, which we do not compare against.
We run experiments on three different tasks, but two of them are low resource and two of them translate into German, which was necessary because we only had access to German language experts for preparing the evaluation.

% \section*{Acknowledgements}

\bibliography{anthology,custom}
\bibliographystyle{acl_natbib}
% Limitations
% \appendix

% \section{Example Appendix}
% \label{sec:appendix}

% This is a section in the appendix.

\clearpage

\appendix

\section{Appendix}
\label{sec:appendix}

\subsection{Pronoun and Formality Translation Evaluation}
\label{subsec:pronoun_formality_eval}
Here, we explain how we calculate the pronoun translation and formality translation F1 scores.

\noindent \textbf{Pronouns} \\
For each triplet ($F_n$, $E_n$, $\hat{E}_n$) (source, hypothesis, reference) of our test data we first check if it contains a valid ambiguous pronoun.
That means, in the source sentence there must be an English 3rd person pronoun in the neutral form and it also must be labeled as a pronoun by the English POS-tagger.
We also check if a 2nd or 3rd person plural pronoun is present in the source and if that is the case, we do not consider female pronouns on the target side, since we could not distinguish if e.g. \lq{}sie\rq{} is the translation of \lq{}it\rq{} or \lq{}they\rq{}.
This would require a word alignment between source and hypothesis/reference which we do not have.
If we found the example to be valid, we then check for occurrences of 3rd person pronouns in the male, female and neuter forms, in both reference and hypothesis using a German POS-tagger as well as language-specific regular expressions.
After going through the complete test data ($F_n$, $E_n$, $\hat{E}_n$) sentence-by-sentence we calculate an F1 score for pronoun translation:
\begin{equation*}
    F1_{pro} = \frac{2 \cdot \text{P}_{pro} \cdot \text{R}_{pro}}{\text{P}_{pro} + \text{R}_{pro}}
\end{equation*}
with precision $\text{P}_{pro} =$
\begin{equation*}
    \frac{\sum_{n=1}^{N} \sum_{x} \text{min}\left(\text{CP}(F_n, E_n, x), \text{CP}(F_n, \hat{E}_n, x)\right)}{\sum_{n} \sum_{x} \text{CP}(F_n, E_n, x)}
\end{equation*}
and recall $\text{R}_{pro} =$
\begin{equation*}
    \frac{\sum_{n=1}^{N} \sum_{x} \text{min}\left(\text{CP}(F_n, E_n, x), \text{CP}(F_n, \hat{E}_n, x)\right)}{\sum_{n} \sum_{x} \text{CP}(F_n, \hat{E}_n, x)}
\end{equation*}
where $\text{CP}(\cdot, \cdot, \cdot)$ counts the number of valid pronoun occurrences and $x \in \{male, female, neuter\}$.

\noindent \textbf{Formality} \\
We follow almost exactly the same steps as for detecting the pronoun translations described above.
The only differences are that we check for validity slightly differently and instead of pronouns we check for occurrences of formal/informal style.
For sentence-pairs where 3rd person female/neuter or 3rd person plural pronouns are present, we do not count the formal occurrences, since we might not be able distinguish the German translations in these cases. 
We calculate an F1 score for formality translation using 
\begin{equation*}
    F1_{for} = \frac{2 \cdot \text{P}_{for} \cdot \text{R}_{for}}{\text{P}_{for} + \text{R}_{for}}
\end{equation*}
with precision $\text{P}_{for} =$
\begin{equation*}
    \frac{\sum_{n=1}^{N} \sum_{x} \text{min}\left(\text{CP}(F_n, E_n, x), \text{CP}(F_n, \hat{E}_n, x)\right)}{\sum_{n} \sum_{x} \text{CP}(F_n, E_n, x)}
\end{equation*}
and recall $\text{R}_{for} =$
\begin{equation*}
    \frac{\sum_{n=1}^{N} \sum_{x} \text{min}\left(\text{CP}(F_n, E_n, x), \text{CP}(F_n, \hat{E}_n, x)\right)}{\sum_{n} \sum_{x} \text{CP}(F_n, \hat{E}_n, x)}
\end{equation*}
where $\text{CP}(\cdot, \cdot, \cdot)$ counts the number of valid pronoun occurrences and $x \in \{formal, informal\}$.

The POS-taggers we use are \texttt{en\_core\_web\_sm}\footnote{\url{https://spacy.io/models/en}} for English and \texttt{de\_core\_news\_sm}\footnote{\url{https://spacy.io/models/de}} for German.
 For both languages, spaCy claims an accuracy of 97\% for POS-tagging and in our testing we did not find even a single error in pronoun-tagging.
For calculating the Pronoun Translation F1 score we use the same ContraPro test set as described in Section \ref{subsec:eval} with the correct references.
For calculating the Formality Translation F1 score, we use the test set from the \textit{OS} En-De task.
The statistics for both test sets are reported in Table \ref{tab:F1_stats}.
\begin{table*}[t]
\centering
\def\arraystretch{1.3}
\begin{tabular}{l|ccc|cc}
\toprule
 & \multicolumn{3}{c|}{Pronoun Trans. F1 score} & \multicolumn{2}{c}{Formality Trans. F1 score} \\ \cline{2-6} 
 & \multicolumn{1}{c|}{neuter} & \multicolumn{1}{c|}{male} & female & \multicolumn{1}{c|}{formal} & informal \\ \hline
\# examples & \multicolumn{1}{r|}{4565} & \multicolumn{1}{r|}{4688} & \multicolumn{1}{r|}{4001} & \multicolumn{1}{r|}{416} & \multicolumn{1}{r}{605} \\ \bottomrule
\end{tabular}
\caption{Number of valid examples for specific ambiguous pronoun/style translation in the reference of our test sets.}
\label{tab:F1_stats}
\end{table*}
In the ContraPro test set, for each gender class we have exactly 4,000 examples.
The fact that we identify more than 4,000 valid examples for the pronoun case means, that in some cases we identify multiple pronouns per sentence.
All in all, we find the classes to be relatively balanced for these test sets.

\subsection{Dataset Statistics and Experimental Setups}
\label{subsec:data_stats}

For the \textbf{NEWS En$\to$De} task, the parallel training data comes from the \texttt{NewsCommentaryV14} corpus\footnote{\url{https://data.statmt.org/news-commentary/v14/}}.
As validation/test set we use the WMT \texttt{newstest2015}/\texttt{newstest2018} test sets from the WMT news translation tasks \cite{farhad2021findings}.
For the \textbf{TED En$\to$It} task, the parallel training data comes from the IWSLT17 Multilingual Task \cite{cettolo2017overview}.
As validation set we use the concatenation of \texttt{IWSLT17.TED.dev2010} and \texttt{IWSLT17.TED.tst2010} and as test set we use \texttt{IWSLT17.TED.tst2017.mltlng}.
For the \textbf{OS En$\to$De} task, the parallel training data comes from the \texttt{OpenSubtitlesV2018} corpus \cite{lison2018opensubtitles2018}.
We use the same train/validation/test splits as \citet{huo2020diving} and additionally remove all segments that are used in the ContraPro test suite \cite{muller2018large} from the training data.
The data statistics for all tasks can be found in Table \ref{tab:data}.

\begin{table}[h!]
\centering
\begin{tabular}{l|l|r|r}
\toprule
\multicolumn{1}{c|}{task} & \multicolumn{1}{c|}{dataset} & \multicolumn{1}{c|}{\# sent.} & \multicolumn{1}{c}{\# doc.} \\ \hline
NEWS & train & 330k & 8.5k \\
 & valid & 2.2k & 81 \\
 & test & 3k & 122 \\
 & ContraPro & 12k & 12k \\ \hline
TED & train & 232k & 1.9k \\
 & valid & 2.5k & 19 \\
 & test & 1.1k & 10 \\ \hline
OS & train & 22.5M & 29.9k \\
 & valid & 3.5k & 5 \\
 & test & 3.8k & 5 \\
 & ContraPro & 12k & 12k \\ \bottomrule
\end{tabular}
\caption{Data statistics for the different document-level translation tasks.}
\label{tab:data}
\end{table}

Since in the original release of ContraPro only left side context is provided, we extract the right side context ourselves from the \texttt{OpenSubtitlesV2018} corpus based on the meta-information of the segments.
For translation of the ContraPro test set, as well as for scoring the contrastive references, we take both the left- and the right-side context into account.
For the full-document systems, we cap the context size for the ContraPro test set to 4 sentences for computational reasons.

We tokenize the data using byte-pair-encoding \cite{sennrich2016neural, DBLP:conf/acl/Kudo18} with 15k joint merge operations (32k for \textit{OS} En$\to$De).
The models are implemented using the fairseq toolkit \cite{ott2019fairseq} following the transformer base architecture \cite{vaswani2017attention} with dropout 0.3 and label-smoothing 0.2 for \textbf{NEWS En$\to$De} and \textbf{TED En$\to$It} and dropout 0.1 and label-smoothing 0.1 for \textbf{OS En$\to$De}.
This resulted in models with ca. 51M parameters for \textit{NEWS} and \textit{TED} and ca. 60M parameters for \textit{OS} for both the sentence-level and the document-level systems.

Let us assume that the training data $\mathcal{C}$ consists of $M$ documents $\mathcal{D}_m$ and each document consists of source-target sentence pairs $(F_{n,m}, E_{n,m})$.
The goal of training is to find the optimal model parameters $\hat{\theta}$ which minimize the loss function:
\begin{equation*}
    \hat{\theta} = \argmin_{\theta} L(\theta)
    \label{eq:training_objective}
\end{equation*}
When training the local context models, we define the loss function:
\begin{equation*}
    L(\theta) = -\frac{1}{|\mathcal{C}|}\sum_{m=1}^{M} \sum_{n=1}^{N_m} \log p_{\theta}(E_{n-k, m}^{n, m} | F_{n-k, m}^{n, m}).
\end{equation*}
When we take full documents as input to the model, the loss function simply becomes
\begin{equation*}
    L(\theta) = -\frac{1}{M}\sum_{m=1}^{M} \log p_{\theta}(E_{1, m}^{N_m, m} | F_{1, m}^{N_m, m}).
    \label{eq:full_doc_train}
\end{equation*}

All systems are trained until the validation perplexity does no longer improve and the best checkpoint is selected using validation perplexity as well.
Training took around 24h for \textit{NEWS} and \textit{TED} and around 96h for \textit{OS} on a single NVIDIA GeForce RTX 2080 Ti graphics card.
Due to computational limitations, we report results only for a single run.
For the generation of segments (see Section \ref{subsec:decoding}), we use beam-search on the token level with beam-size 12 and length normalization.

\end{document}